\documentclass{article}
\usepackage{authblk}
\usepackage{rotating}
\usepackage{graphicx}
\usepackage{tabularx} 
\usepackage{booktabs} 

\usepackage{multirow}
\usepackage{adjustbox}
\usepackage{algorithmic}
\usepackage{xurl}
\usepackage{amsmath,amssymb,amsfonts}
\usepackage[ruled, noend]{algorithm2e}
\SetKwComment{Comment}{$\triangleright$\ }{}
\newcommand{\B}{\fontseries{b}\selectfont}
\mathchardef\mhyphen="2D 
\usepackage{xcolor}
\newcommand{\ignore}[1]{}
\begin{document}
\title{Run-of-Mine Stockyard Recovery Scheduling and Optimisation for Multiple Reclaimers}
\author[1]{Hirad Assimi}
\author[2]{Ben Koch}
\author[2]{Chris Garcia}
\author[1]{Markus Wagner}
\author[1]{Frank Neumann}
\affil[1]{School of Computer Science, The University of Adelaide, Adelaide, 5000, South Australia, Australia}
\affil[2]{EKA Software Solutions
Adelaide, South Australia, Australia}

\makeatletter
\renewcommand\AB@affilsepx{, \protect\Affilfont}
\makeatother

\affil[1]{firstname.lastname@adelaide.edu.au}
\affil[2]{firstname.lastname@ekaplus.com}

\setcounter{Maxaffil}{0}
\renewcommand\Affilfont{\itshape\small}

\date{}
\maketitle
\begin{abstract}
Stockpiles are essential in the mining value chain, assisting in maximising value and production. Quality control of taken minerals from the stockpiles is a major concern for stockpile managers where failure to meet some requirements can lead to losing money. This problem was recently investigated using a single reclaimer, and basic assumptions. This study extends the approach to consider multiple reclaimers in preparing for short and long-term deliveries. The engagement of multiple reclaimers complicates the problem in terms of their interaction in preparing a delivery simultaneously and safety distancing of reclaimers. We also consider more realistic settings, such as handling different minerals with different types of reclaimers. We propose methods that construct a solution step by step to meet precedence constraints for all reclaimers in the stockyard. We study various instances of the problem using greedy algorithms, Ant Colony Optimisation (ACO), and propose an integrated local search method determining an efficient schedule. We fine-tune and compare the algorithms and show that the ACO combined with local search can yield efficient solutions.
\end{abstract}

\section{Introduction}  
Evolutionary algorithms (EAs) have been applied to challenging combinatorial and continuous optimisation problems in real-world applications of mining. EAs, for example, have been employed in the iron mine supply chain optimisation for long-term planning of mine digger equipment and trucks for commercial purposes~\cite{ibrahimov2014scheduling}. Ant colony optimisation~\cite{Dorgio2004} has been applied to solve the open-pit mining scheduling problem~\cite{shishvan2015long}. Recently, differential evolution~\cite{storn1997differential} has been used to tackle the stockpile blending problem considering the average grade of stockpiles and addressing a continuous optimisation problem subject to uncertainty~\cite{YueCEC,YueGecco}.

The extraction of economically valuable minerals in the form of ore is a part of mine production. Haul trucks transport and stack the mined ore into Run-of-Mine (ROM) stockpiles. As the next step in the value chain to prepare a \textit{delivery} to the client, stockpile schedulers select a blend of the material in ROM stockpiles to be sent to the crushers for processing. We refer to this blend planning as the \textit{stockpile recovery} scheduling. Quality control of deliveries is a major concern, as higher-grade ore is frequently blended with lower-grade ore to guarantee that the delivery matches the customer's mineral needs. However, the selection is difficult due to the complexity of the ROM stockyard deposits, such as various minerals irregularities. 

Stockpile schedulers aim to figure out how to plan stockpile recovery to keep upcoming deliveries qualities consistent in terms of economic minerals, meeting clients' requirements and lowering operation costs. In practice, human specialists plan the stockyard recovery using the rules of thumb based on available data from laboratory samples and a rough estimate of the average grade of ore in the stockpile. However, this is an error-prone decision-making process with insufficient decision support with respect to the technical challenges making it difficult to account for planning a long term stockpile recovery. 
A poor stockpile recovery plan can have serious consequences. Substantial penalty fees, a lack of consistency in decision-making and operations, increasing operating costs, unanticipated losses in practice, and fluctuations in the assumed profit can all be consequences.

Developing an effective recovery plan is crucial to improve decision-making for stockyard recovery and to consider the above-mentioned objectives, correct human shortcomings, and enable long-term delivery planning.
\subsection{Related Work} expanded
Stockpiles are critical components in mining supply chains since they potentially increase net present value by maintaining a mining buffer. 
Examples include ROM stockpiles in copper production and dry bulk terminals in iron and coal production. Reclaimer scheduling in dry bulk terminals has been well explored~\cite{angelelli2016reclaimer,unsal2019exact}, where they are mainly a variation of the parallel machine scheduling problem. However, the main shortcoming in these studies is that they adopt a supply chain model that considers the stockpiles as a whole. However, different cuts in a stockpile can have varying tonnage and mineral compositions in practice.

To mitigate this issue, Lu and Myo~\cite{lu2010optimization,lu2011optimal} investigated stockpile recovery optimisation while considering a stockpile discretised into cuts.  Their model aims to minimise a reclaimer's energy consumption. They calculate the relocation cost of a machine by Euclidean distance from one place in the stockyard to another. They represent their problem as a mixed-integer programming model dealing with a small-scale problem considering two stockpiles and two upcoming deliveries with a single reclaimer.

Their approach is limited in adapting to a real-world case: (1) relocation expenses calculation is more complex than a simple Euclidean distance; (2) the stockyard in practice is bigger, and the number of cuts with respect to the stockpile visibility resolution can determine the size and complexity of an optimisation problem that has been underestimated; (3) another key factor is the number of upcoming deliveries that can be incorporated in the schedule; (4) They only consider bucket-wheel-reclaimer; however other types of reclaimer machines can be used in stockyard recovery. (5) reaching the target quality grade of delivery encompasses more technical features than a weighted average penalty.

In a recent paper, Assimi et al. modelled the stockpile recovery scheduling problem with considering more realistic settings than previous studies~\cite{SAC21}. They defined the problem as a combinatorial optimisation problem with precedence constraints to address the inaccessibility issue of the cuts in the stockyard. They introduced an objective function to prioritise penalty fees over operation costs with respect to the end-user preferences.

The study's main disadvantage is that it only considers a single reclaimer machine in the stockyard; however, stockyards are reclaimed in parallel using multiple reclaimers in practice to prepare deliveries. Another flaw is that they only consider a single type of reclaimer with a single mechanical direction of reclamation. However, in practice, the types of reclaimers can be different, and the variety can bring up different reclaiming directions using their mechanical arms. They consider one row, and four stockpiles of a single type of material, along with preparing for up to four deliveries. Despite its shortcomings, they used solution construction based methods such as greedy algorithms and ant colony optimisation with local search to build a valid solution step by step. They showed that deterministic greedy algorithms might fail to obtain good solutions when the number of upcoming deliveries grows. However, ant colony optimisation with local search can be a viable solver to these complex problems.

There is currently a lack in the literature for dealing with stockpiles in more realistic scenarios. When multiple reclaimers are present, the reclamation direction can shift, different types of material must be delivered, and the stockpile manager should deal with longer-term delivery plans.
\subsection{Our Contribution} 

In this paper, we investigate the stockpile recovery scheduling problem using multiple reclaimers with more realistic settings than before. The stockyard is defined as a directed graph with cuts as vertices and direction as the mechanical \textit{reclaiming direction}. The reclaiming direction is an inherent element of a reclaimer machine; for example, front end loaders  can only reclaim in one way, while bucket wheel reclaimers can reclaim in two directions.

Our industry partner can provide visibility inside a stockyard by using GPS data on dumpsites and laser scanning. As a result, a large stockpile, commonly considered a whole, can be divided into smaller cuts. Our mission is to obtain an efficient plan for multiple reclaimers preparing deliveries sequentially. We consider the problem regardless of stockyard economic minerals or reclaimer machines to reach a scalable method.

To avoid penalty fees and reduce operational costs, we aspire to achieve a good delivery schedule for each reclaimer while preserving the target quality of deliveries. 
The obtained schedules refer to the sequence in which each reclaimer reclaims cuts at a given time step according to the reclaimers' reclaiming directions.

Precedence relationships between cuts limit the search space by restricting how cuts can be accessed regarding their position in the stockyard. To guarantee that the precedence restrictions are met in a final solution, we must ensure that every schedule segment is valid. Therefore we use solution construction heuristic approaches that can construct a solution step by step. We also need a method that allows us to deal with real-world problems that address different objective functions and constraints and effectively incorporate additional technological limitations based on end-user requirements for future additions.

These methods require a solution construction heuristic. In this study, to simulate the interaction among reclaimers and ensure the safety distancing, we develop a custom solution construction heuristic to consider all reclaimers in preparing a delivery simultaneously and sequentially. To select a job at each step, we employ selection strategies from deterministic and randomised greedy algorithms, indicated as DGA and RGA, respectively, and Max-Min Ant System (MMAS) with a customised local search to tackle the problem. Our experiments consider realistic stockpile recovery settings integrating different types of reclaimers, material, and interaction among the reclaimers. The source code is available at \url{https://git.io/JyUI1}.

The rest of the paper is organised as follows. In the following section, we define the stockpile recovery problem using multiple reclaimers before explaining the objective function. Next, we present the optimisation algorithms and solution construction heuristic for simulating the reclaimers interactions. Then, we set up experiments, fine-tune our algorithms and report on the behaviour and quality of obtained solutions for each problem instance. We find that MMAS with local search can outperform other methods in most of the case studies.

\section{Problem Statement} 
 
In this section, we define the stockpile recovery scheduling optimisation using multiple reclaimers in the form of a combinatorial problem.

\subsection{Stockyard Model}
  
We model a given stockyard in form of a directed graph $\mathcal{G} = (\mathcal{C}, \mathcal{E})$ without cycles where $\mathcal{C}=\{c_1, c_2, \dots, c_J\}$ is the set of cuts in the stockyard including $J$ cuts with index $j$.  Each cut denotes a slice of a stockpile at a certain location in the stockyard with respect to the positioning of its bench, stockpile and row in the stockyard.
 
Each cut contains information on the mineral compositions of economic and contaminant minerals, denoted by $m_{j,k}$ where $k = \{1,2, \dots, \mathcal{K}\}$, which defines the set of chemical elements preferred by the end-user to be evaluated.

Stockpile recovery is challenging due to technical restrictions, including chemical concentrations and operational constraints. Chemical concentration restriction ensures the grade quality of the delivery where undesired chemical contaminants in a delivery fall within the customer's permitted range. If a delivery fails to fulfil this restriction, the stockpile manager should pay financial penalties proportionate to the severity of the violation.
Operational constraints include the accessibility of cuts in the stockyard where the lower benches are unreachable before reclaiming the higher bench, and reclaimers should distance themselves safely to avoid collisions.

In the stockyard, we assume the stockpiles are parallel to each other. This assumption allows us to consider several types of reclaimers. Some reclaimers, such as bucket wheel reclaimers, can move their mechanical arm in two different directions during recovery, which we refer to as reclaiming direction ($\phi$). Front-end loaders, on the other hand, can only move their arms in one direction. 
 
Each cut $(c_j)$ has a specified tonnage given by $\Gamma_j$, and reclaiming a cut takes a specific amount of time proportionate to its size denoted by $\tau_{j}$. $\mathcal{E}$ denotes the set of edges representing the immediate predecessors for each cut with its corresponding reclaiming direction. $e_{j,\phi_j} \in \mathcal{E}$ denotes the edge connecting to the $c_j$ with reclaiming direction of $\phi_j$; the edge physically represents the reclaimer \textit{job} where reclaimer relocates from its current position to $c_j$ and reclaims the destination cut with reclaiming direction of $\phi_j$. Precedence constraints determine the validity of a schedule. If it fails for a segment of the solution, the schedule becomes unable to be processed.
 
\begin{figure}[t]
    \centering
	\includegraphics[width=0.8\textwidth]{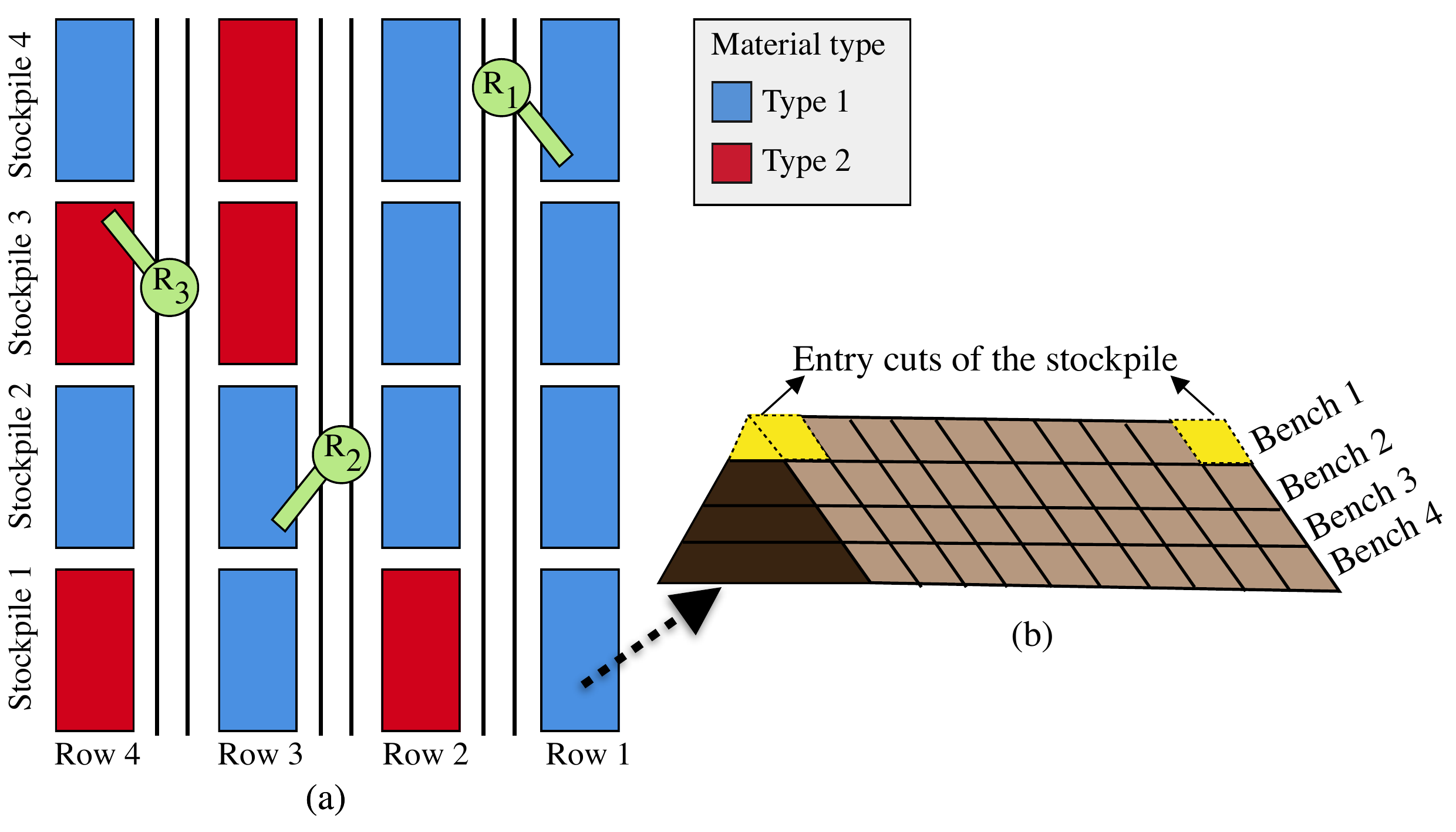}
	\caption{(a) Top view of the stockyard configuration (b) Layout of a single stockpile with four benches each containing ten cuts}
	\label{fig.stockyard}
\end{figure}

Figure \ref{fig.stockyard} shows the stockyard configuration in our problem. We have four rows, each one containing four stockpiles. Stockpiles can be split into 4 benches, each of which has 10 cuts. We suppose there are two kinds of deliveries in terms of the particle size distribution of stacked material. Each stockpile can only contain one type of material. Entry cuts refer to the first cut of the stockpile that can be accessed with a reclaimer with a specific reclaiming direction. At most, three reclaimers are parallel to each other.
 
Stockpile recovery is a time-consuming operation that entails the relocation of reclaimers, such as front-end loaders and bucket wheel reclaimers, in the stockyard and the time it takes to reclaim a cut proportionate to its size. A cut is a section of piled material that can weigh 1000 to 5000 tonnes. Operation time and operation costs are phrases that can be used interchangeably. To calculate a reclaimer's relocation cost shown as $\mathcal{T}_{i, j, \phi_i, \phi_j}$, we need to know its current position at cut $i$, the destination ($j$), and their corresponding reclaiming directions, $\phi_i$ and $\phi_j$, respectively.
   
A candidate solution for a schedule, $x$ represents how jobs are scheduled for reclaimers. Sets $R$ and $D$ denote the reclaimer and delivery sets, respectively. We show a sub-segment of $x$ for reclaimer $r$ in preparation of delivery $d$ as $x_d^r$.

We aim to maintain the target quality of deliveries in terms of economic and contaminant minerals while lowering operating costs based on end-user preferences. The first objective is to guarantee that the average of chemical minerals in delivery remains within a predefined range depending on the type of material in delivery. We calculate the average target quality for each mineral ($k$) with respect to the delivery $d$ as follows. Note that to form $x_{d,k}$ we consider all reclaimers.
$$\hat{x}_{d,k} = \frac{1}{|x_d|} \sum_{e_{j,\phi_j} \in x_d} m_{j,k}$$
where, the predefined range constraint is
$$\underline{m_{d,k}} \leq \hat{x}_{d,k} \leq \overline{m_{d,k}} \quad \forall k \in K$$
Each mineral has a different magnitude, as well as a lower and maximum limit. As shown below, we utilise the bracket-operator penalty approach~\cite{deb2001multi} to evaluate the degree of violation for \textit{average target quality}.
$$\big\langle \hat{x}_{d,k} \big\rangle = \left \{
\begin{array}{ll}
\frac{|\hat{x}_{d,k}-\overline{m_{d,k}}|}{|\overline{m_{d,k}}|} & \text{if}\; \hat{x}_{d,k} > \overline{m_{d,k}}\\
\frac{|\hat{x}_{d,k}-\underline{m_{d,k}}|}{|\underline{m_{d,k}}|} & \text{if}\; \hat{x}_{d,k} < \underline{m_{d,k}}\\
0 &  \underline{m_{d,k}} \leq \hat{x}_{d,k} \leq \overline{m_{d,k}}\\
\end{array}
\right.$$
To calculate the total violation for average target quality in $x$ for all reclaimers for the delivery $d$ we have:
$$v_1(x_d) = \sum_{k=1}^{K} \big \langle \hat{x}_{d,k} \big\rangle$$ 

Another objective is to ensure that a massive delivery in different packages remains consistent. We refer to it as \textit{window target quality}.

$$\widetilde{x}_{d,j,k} =  \frac{m_{j,k}+m_{j-1,k}+m_{j-2,k}}{3}$$

We use the bracket operator, as we did for the first objective, to determine the degree of violation for this objective, and we have:
$$v_2(x_d) = \sum_{k=1}^{K}  \sum_{j=4}^{|x_{d,k}|} \big\langle \widetilde{x}_{d,j,k} \big\rangle$$ 

$j$ refers to the position of a cut in the solution segment since we want to calculate the window target quality after three cuts have already been reclaimed for a delivery. The third objective is to lowering the operation costs, where we want to have a schedule for reclaimers to prepare the deliveries faster. We define a utility function as follows.
$$ u(x_d) = \sum_{c_j \in x_d} \frac{T_{e_{j,\phi_j}}}{\Gamma_{j}} $$

where $T_{e_{j,\phi_j}}$ is the cost required to complete job $e_{j,\phi_j}$ considering the cost of relocation and cut reclamation.
$$ T_{e_{j,\phi_j}} = \mathcal{T}_{i, j, \phi_i, \phi_j} + \tau_{j}$$

\subsection{Objective Function} \label{sec.obj}
We prioritise target quality over operational costs because if the target quality is violated, the stockpile manager must pay financial penalties. As a result, the primary objective is to avoid violating the average target quality; the second priority is to avoid violating window target quality and, subsequently, to reduce the utility. We employ a lexicographic objective function for minimisation to consider these priorities.
$$f(x)=\left(\sum_{d=1}^{D} v_1(x_d), \sum_{d=1}^{D} v_2(x_d), \sum_{d=1}^{D} u(x_d)\right)$$

Order in the objective function matters. To compare two solutions $x$ and $z$, we have:
\begin{align*}
	f(x) &\leq f(z) \\
	\text{iff} \quad v_1(x) &\leq v_1(z) \lor \\
	(v_1(x) &= v_1(z) \land v_2(x) \leq v_2(z)) \lor \\
	(v_1(x) &= v_1(z) \land v_2(x) = v_2(z) \land u(x) \leq u(z))
\end{align*}

For example, among $f(x)=(0.01, 0, 35)$ and $f(z)=(0, 0.01, 15)$ and $f(w)=(0, 0, 45)$, first, we look at the first component, and we see that $v_1$ is zero for solutions $z$ and $w$; thus, solution $x$ is the worst of all solutions. To compare solutions $z$ and $w$, we consider the second component ($v_2$) where solution $w$ outperforms the solution $z$. We have:
$$ f(w) < f(z) < f(x) $$

Our problem can be viewed as an extended version of the Travelling Salesperson Problem (TSP) with real-world constraints and objectives. In our case, we replace the cities and TSP distances with mineral cuts and the reclamation job costs, respectively. Furthermore, we have several agents (reclaimers) on graphs that must adhere to a no-cross condition.  We aim to find a solution to maintain the objectives with respect to the end-user preferences. There are precedence constraints in accessing the cuts and requirements to maintain the target quality of deliveries.

\section{Optimisation Methods}
 In this section, we explain our methods. First, we present the solution construction heuristic, which simulates the reclaimers interactions to construct a schedule. Next, we explain the employed algorithms that generate a valid solution step by step while adhering to the precedence constraints. We look at deterministic and randomised variants of greedy algorithms and ant colony optimisation with and without the local search.
 
\subsection{Solution Construction Heuristic}
Algorithm \ref{alg.SCH} shows the solution construction heuristic procedure.We assume that each reclaimer can only do one reclamation job at a time. Reclaimers cannot be interrupted while performing a reclaiming job, and the job must be completed before starting another (non-preemptive constraint).

Each reclaimer can only reclaim from its adjacent stockpiles. For example, in Figure \ref{fig.stockyard}, $R_1$ can only reclaim from Row 1 and Row 2, and all stockpiles in Row 2 are shared between $R_1$ and $R_2$. However, reclaimers on a shared row can not get too close to one another in terms of a safety distance constraint. 

Reclaimers can be \textit{idle} or \textit{busy} at a time. Initially, the solution set $x$ is empty, and reclaimers are idle. Reclaimers have access only to the entry cuts at their adjacent stockpiles. The entry cuts are positioned opposite with respect to the other reclaiming direction.

We assume that the reclaimers begin with reclaiming the fixed entry cut with the fixed initial direction ($\phi_0$) from a stockpile with the material type of first delivery ($d=1$). We denote the initial reclamation job as $e_{c_0^r,\phi_0}$ where $c_0^r$ is the fixed entry cut for reclaimer $r$. Reclaimers record a completion time of the job as they become \textit{busy} and we push the cut to their queue.
When the initial job's completion time has passed, the reclaimer(s) become idle, and we add the reclaimed cut to the reclaimer schedule. Next, we can figure out what cuts are accessible for a reclaimer. $\mathcal{N}_t^r$ shows the available cuts in the neighbourhood of reclaimer $r$ at its position at time step $t$. If more than one reclaimer becomes idle at the same time, the total neighbourhood is a set containing candidates from each reclaimer paired.
Note that each reclaimer's neighbourhood is made up of a variety of reclaiming directions. The following criteria are used to choose eligible cuts for inclusion in $\mathcal{N}_t$.
\begin{itemize}
	\item Reclaimers should not get too close to one another in terms of a safety distance
	\item The deliveries are prepared sequentially. However, suppose a reclaimer exhausts all cuts of a specific type of delivery. In that case, for idle reclaimer, we begin reclaiming the following delivery (with a different type of mineral) to save time and be more efficient. In our settings, we refer to this exception as \textit{material exhaustion exception}. 
\end{itemize}

Next, the idle reclaimer(s) select a job from its eligible neighbourhood as $e_{j,\phi_j}\in \mathcal{N}_t^r$. We evaluate the quality of each job as described in Section \ref{sec.obj}, and the selection is carried out according to the applied algorithm selection approach. This procedure is repeated for all reclaimers, and after completion of each job, the stockyard graph is updated. The reclaimers continue to reclaim minerals until we reach the specified tonnage of delivery $d$. The reclaimer will then begin reclaiming the next delivery ($d+1$). Note that if multiple reclaimers are idle at the same time, we evaluate a combination of jobs with respect to the reclaimers. This cycle is repeated for each delivery until all deliveries are completed. 
\begin{algorithm}[t]
\caption{Solution Construction Heuristic}\label{alg.SCH}
\DontPrintSemicolon
\SetKwInOut{Input}{Input}
\SetKwInOut{Output}{Output}
\Input{Selection algorithm from [DGA, RGA, MMAS]}
$t \gets 0$\;
$x := \emptyset$ \Comment*[l]{Let $x$ be the solution set}
$ \text{job}^r \gets e_{c_0^r},\phi_0 \quad \forall r \in R$
\Comment*[l]{Add initial job to the queue}
\ForEach{$r \in R$}{
$d^r \gets 1$\;
$\text{status}^r \gets \textit{busy}$\;
$t^r \gets T_{\text{job}^r}$ \Comment*[r]{Record completion time}
}
\While{$d \leq D$  \Comment*[r]{all deliveries are not planned}}
{
$t \gets t+1$ \Comment*[r]{next time step}
\ForEach{$r \in R$}{
\If{$t^r \leq t$}{
$x_d^r \gets  \text{job}^r$ \Comment*[r]{Add completed job to $x$}
$ \Gamma_{x_d} = \Gamma_{x_d} + \Gamma_{\text{job}^r}$ \Comment*[r]{Tracking tonnage of the delivery}
\If{$\Gamma_{x_d} > \Gamma_d$}{
$d = d +1$  \Comment*[r]{We start reclaiming the next delivery}
}
Generate $\mathcal{N}^r_t$  \Comment*[r]{wrt. $d^r$}
\While{${N}^r_t = \emptyset \land d^r <= D$}
{$d^r = d^r + 1$ \Comment*[r]{material exhaustion exception}
Generate $\mathcal{N}^r_t$ \;
} 
} 
} 
$\displaystyle \mathcal{N}_t = \prod_{r}^{R} N_t^r \; \forall r \in R$ if $r$ is idle and $\text{type}_d^r$ is identical  \;
\textbf{Evaluation} of all candidate jobs in $\mathcal{N}_t$\;
\textbf{Selection} of next job wrt. to input algorithm probability of selection
}
\end{algorithm}
\subsection{Deterministic and Randomised Greedy Algorithm}
\begin{algorithm}
\caption{Greedy Algorithms}\label{alg.GA}
\DontPrintSemicolon
$x := \emptyset$ \;
Construct a solution $x$ step by step, using Algorithm \ref{alg.SCH} with following selection probability\;
\If{Selection is deterministic (DGA)}{
Choose a job $e^*=\operatorname*{argmin}_{e_{j,\phi_j} \in \mathcal{N}_t} f(e_{j,\phi_j})$\; 
}
\Else { \Comment*[l]{RGA}
Choose a job $(e^*)$ according to probability $p(e_{j,\phi_j} | \mathcal{N}_t) = \dfrac{\eta(e_{j,\phi_j})^\lambda}{\sum_{e_{l,\phi_l} \in \mathcal{N}_t} \eta(e_{l,\phi_l})^\lambda}$ \;
}
\end{algorithm}

Greedy algorithms are simple to use and can tackle large combinatorial problems quickly. However, there is no guarantee of success because they may get trapped in local optima. Algorithm \ref{alg.GA} shows the procedure for the deterministic and randomised greedy algorithm, shown as DGA and RGA, respectively.

DGA selects the best eligible job with the highest greediness at each time step as follows.
$$e^* = \operatorname*{argmin}_{e_{j,\phi_j} \in \mathcal{N}_t} f(e_{j,\phi_j})$$
In the scheduling process, acting deterministic greedy can result in reclaiming the best cuts in the stockyard early. If all high-quality cuts in the stockyard are exhausted too soon, DGA may fall into the trap of local optima and fail to plan the next deliveries effectively~\cite{SAC21}. Other studies such as~\cite{gao2018randomized} recommend controlling DGA greediness to enhance its efficiency. The randomised greedy algorithm (RGA) is a basic variant of the DGA that can help us find better solutions by controlling greediness.

RGA has a greedy control parameter $\lambda \geq 0$. When $\lambda$ is small, RGA gives weight to candidates in the selection where there is still a chance for weaker candidates to be selected; however, when $\lambda \to \infty$, RGA behaves similarly to DGA.

The probability of selection of a cut in a neighbourhood of reclaimers at time step $t$ for RGA is:
$$p(e_{j,\phi_j} | \mathcal{N}_t) = \dfrac{\eta(e_{j,\phi_j})^\lambda}{\sum_{e_{l,\phi_l} \in \mathcal{N}_t} \eta(e_{l,\phi_l})^\lambda}$$
This probabilistic selection is analogous to a roulette wheel, with the probability of selecting a candidate in $\mathcal{N}_t$ is proportionate to $\eta(e_{j,\phi_j})^\lambda$.
$\eta$ refers to the heuristic information, and we calculate it as follows. 

\begin{figure}[t] 
    \centering
	\includegraphics[width=0.75\textwidth]{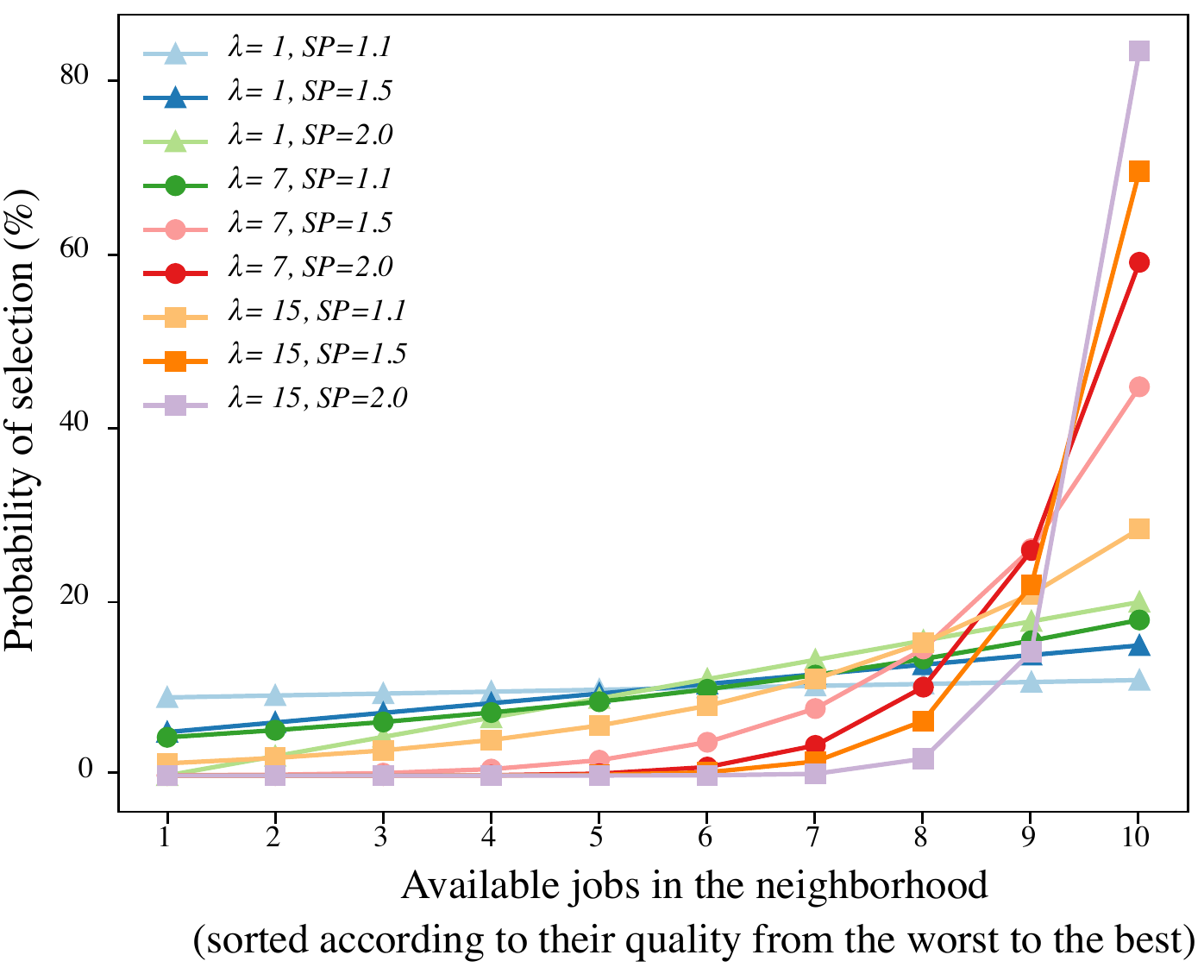}
	\caption{Probability of selection when linear ranking is active for different values of $\lambda$ and $SP$}
	\label{fig.sp}
\end{figure}

If the average target quality and window target quality for all candidates are zero (
$v_1(e_{j,\phi_j}) = 0 \land v_2(e_{j,\phi_j}) = 0 \quad \forall e_{j,\phi_j} \in \mathcal{N}_t$), 
$\eta(e_{j,\phi_j})$ will be as identical as $u(e_{j,\phi_j})$. Otherwise, we employ linear ranking as follows.
We sort the jobs from the best to the worst with index $i$ from 0 to $|\mathcal{N}_t|-1$.
We calculate the probability of selection based on their rank as follows~\cite{Eiben2015}.
$$\eta_{e_{i,\phi_i}} = \dfrac{2-SP}{\mu} + \dfrac{2i(SP-1)}{\mu(\mu-1)}$$
Figure \ref{fig.sp} shows the behaviour of the preceding equation with respect to different values of $\lambda$ and $1< SP \leq 2$ where the latter refers to the selection pressure parameter.

\subsection{Max-Min Ant System (MMAS)}
Ants in nature take a random walk from their nest to locate a food source. On their way back to the nest, ants leave a substance called pheromone. Other ants can detect it and follow the favourable path for food. As more ants go along this path, the pheromone becomes stronger and more ants follow the favourable path. However, some pheromone progressively evaporates over time, resulting in reducing the attractiveness capability of unexplored routes.

Ant Colony Optimisation (ACO)~\cite{Dorgio2004} represents a mathematical platform employing artificial ants based on the foraging behaviour of ants. The probabilistic selection is based on pheromone and heuristic information. Pheromone aid in transferring knowledge for future generations, and heuristic information serves as a quality indicator of the path with respect to our objective function. For our study, ACO can be seen as an iterative RGA in which the pheromone plays a part in selection jobs to construct a solution.
\begin{algorithm}[t]
\caption{Max-Min Ant System (MMAS)}\label{alg.MMAS}
\DontPrintSemicolon
\textbf{initialise} $\xi$ \Comment*[r]{pheromone matrix initialisation} 
Generate $n$ ants for initial colony $\pi_i \subset \Pi$ \;
\Repeat{MMAS termination criterion met}{
\For {each ant $\pi_i$}{\Repeat{solution is complete}{construct a solution $x$ step by step, using Algorithm \ref{alg.SCH} with MMAS selection probability}}
Update best found solution\;
Update $\xi$\;
}
\Return best found solution
\end{algorithm}

MMAS~\cite{STUTZLE2000889} is a well-known variant of ACO which has been previously used for different applications~\cite{chagas2020ants}. Algorithm \ref{alg.MMAS} shows the procedure for MMAS. MMAS sets the initial value for the pheromone matrix for all edges (all possible jobs) to $\xi_0$. Next, MMAS generates $n$ artificial ants for its colony at the first generation. Each ant performs a random walk using the solution construction heuristic to generate a valid solution step by step. employing the following probabilistic selection strategy. 
$$p(e_{j,\phi_j} | \mathcal{N}_t) = \dfrac{[\xi_{e_{j,\phi_j}}]^\alpha [\eta(e_{j,\phi_j})]^\beta}{\sum_{e_{l,\phi_l} \in \mathcal{N}_t}[\xi_{e_{l,\phi_l}}]^\alpha [\eta(e_{l,\phi_l})]^\beta}$$
where $\alpha$ and $\beta$ are MMAS parameters that regulate the influence of heuristic and pheromone information, respectively, the preceding equation implies that the selection of the next cut is both dependent on the quality of jobs in terms of the objective function and the pheromone information. MMAS restricts $\xi_{e_{j,\phi_j}}$ in the range $\left[\dfrac{1}{J}, 1-\dfrac{1}{J}\right]$~\cite{neumann2009analysis}.

After all ants complete their random walk, pheromone evaporation occurs where it aids in avoiding the unvisited paths as follows.
$$\xi_{e_{j,\phi_j}} = (1-\rho).\xi_{e_{j,\phi_j}}$$
where $0 < \rho \leq 1$ denotes the evaporation rate. Then, one specified ant ($\pi^*$) deposits pheromone on their solution ($x^*$) edges as follows.
$$\xi_{i,j,\phi_i,\phi_j} = \left \{
\begin{array}{ll}
   \text{min}\{\xi_{e_{j,\phi_j}}+ \rho, 1-\dfrac{1}{J}\} & \text{ if } (e_{j,\phi_j}) \in x^* \\ 
   \text{max}\{\xi_{e_{j,\phi_j}}, \dfrac{1}{J}\} & \text{otherwise.}
\end{array} \right.$$
$\pi^*$ could be the best-so-far-ant (BSFA) or the iteration-best-ant (IBA). 
\subsection{Iterative Local Search}
\begin{algorithm}[t]
\caption{Iterative Local Search for MMAS}\label{alg.ILS}
\SetKwInOut{Input}{Input}
\DontPrintSemicolon
\Input{x = Initial Solution}
Stop criterion $\gets$ False\;
\While{Stop criterion $=$ False}{
\ForEach{$d \in D$}{
\ForEach{$r \in R$}{
Successful swap $\gets \emptyset$\;
\ForEach{$e_{j,\phi_j} \in x_d^r$}{
x* = Swap(x) \Comment*[r]{Swap this job with its succeeding adjacent} 
\If{precedence constraints in $x^*$ are met}{
Calculate new completion time for affected jobs\;
Calculate $u(x^*)$ and $v_2(x^*)$\;
\If{$f(x^*) < f(x)$}{
Record $x^*$ as a successful swap\;
}
}
}
$ x \gets$ Best solution out of recorded solutions in successful swaps\;
}
}
}
\Return x
\end{algorithm}

We apply iterative local search (ILS)~\cite{lourencco2003iterated} to improve solutions obtained by MMAS. ILS can search the neighbourhood of an obtained solution finely. Algorithm~\ref{alg.ILS} shows the procedure of the ILS for our problem.

It is hard to define a local search neighbourhood for an obtained schedule produced for multiple reclaimers which each delivery is scheduled sequentially. The precedence connection between different reclaimers with respect to shared rows can complicate the issue. As a result, a small neighbourhood is defined as follows. We consider the deliveries in order and we begin with the first reclaimer, as $x_{d=1}^{r=1}$.

In this segment, we can swap the positions of two adjacent jobs to get the perturbed solution as $x^*$. Next, we validate the precedence constraints with respect to all reclaimers. If it meets the constraints, we evaluate the change in the objective function. Note that swapping adjacent jobs in a the defined segment only affects $v_2(x)$ and $u(x)$, but $v_1(x)$ remains unchanged.
Due to this neighbourhood, $v_1(x)$ does not change and only $v_2(x)$ and $u(x)$ are affected.
If the perturbed solution outperforms the original solution, we mark it as a successful swap, and  if necessary, we reorder the jobs to ensure time consistency in terms of affected jobs' completion time. 

After swapping all adjacent jobs in $x_{d=1}^{r=1}$, the best one from the recorded ones replaces the original solution, and we repeat the local search on the same segment. The process is repeated until no improvement is observed. Then we proceed to the next segment as the next reclaimer for the same delivery, and the iterative local search technique described above is repeated. 

\section{experimental setup} 
In this section, we detail our experimental setup and assumptions and fine-tune the algorithms we described in the previous section.

\subsection{Problem Setup}
We consider a real-world discretised stockyard model provided by our industrial partner shown in Fig~\ref{fig.stockyard} that described previously. We suppose that no more than ten deliveries should be planned (according to available material in the stockpile), with each delivery requiring a tonnage in range of [100,000, 200,000] tonnes. Note that the stockyard dataset allows us to schedule up to 10 deliveries, whereas more than 9 deliveries can be processed only with 3 reclaimers. 

We define an instance as a three-tuple ($R, D,\phi$) with the following components in order: number of deliveries, number of reclaimers and how many reclaiming direction is possible. For example, (6-2-2) depicts a situation in which six deliveries should be planned, two reclaimers are available, and reclaimers can employ both $\phi_1$ and $\phi_2$ reclaiming directions. In all instances, we evaluate the objective function in the form $(v_1(x), v_2(x), u(x))$.

 \subsection{Algorithm Setup}
We see that our algorithms have several parameters that can influence the computational cost and their behaviour to discover good solutions. 

DGA is deterministic, and there is no configuration parameter. On the other hand, RGA has two parameters $\lambda$ and $SP$ to adjust greediness and selection pressure, respectively. We use $\lambda \in \{1,2,3, \dots, 15\}$ and $SP \in \{1.1,1.2,1.3,...,2\}$.

For MMAS, we set the number of ants as 10 and the termination criterion as 1000 maximum generations since they affect the computational expense. MMAS have parameters of $\alpha$, $\beta$, $\rho$, $n$, and $\pi^*$ to be configured as follows.
 
 \begin{itemize}
 \item $\alpha \in \{1,2,3 \dots, 10\}$
 \item $\beta \in \{1,2,3 \dots, 10\}$
 \item $\rho \in \{0.1,0.2,0.3, \dots, 1.0\}$
 \item $SP \in \{1.1,1.2,1.3,\dots,2.0\}$
 \item $\pi^* \in \{BSFA, IBA\}$.
 \end{itemize}
 
To find a suitable parameter configuration, we use the Irace software package~\cite{irace} which employs the method of 1/F-Race~\cite{frace} for automatic algorithm configuration. We use the default parameters and limit the number of experiments to 5000 as the termination criterion for parameter tuning.

\begin{figure}[t] 
	\centering
	\includegraphics[width=0.75\linewidth]{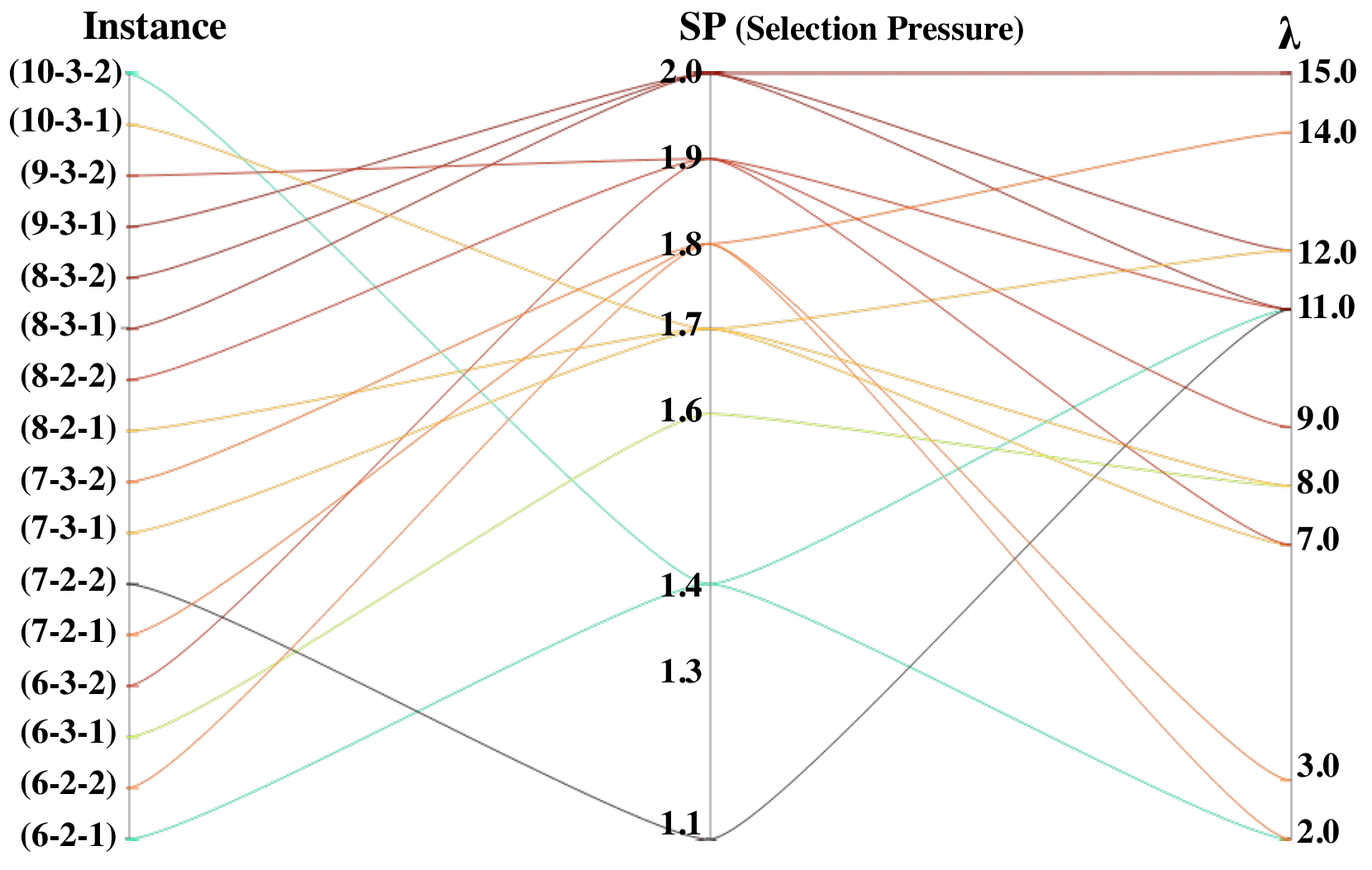}
	\caption{Best parameter configurations for RGA}
	\label{fig.irace_rga}
\end{figure}

\begin{figure}[t] 
	\centering
	\includegraphics[width=0.75\linewidth]{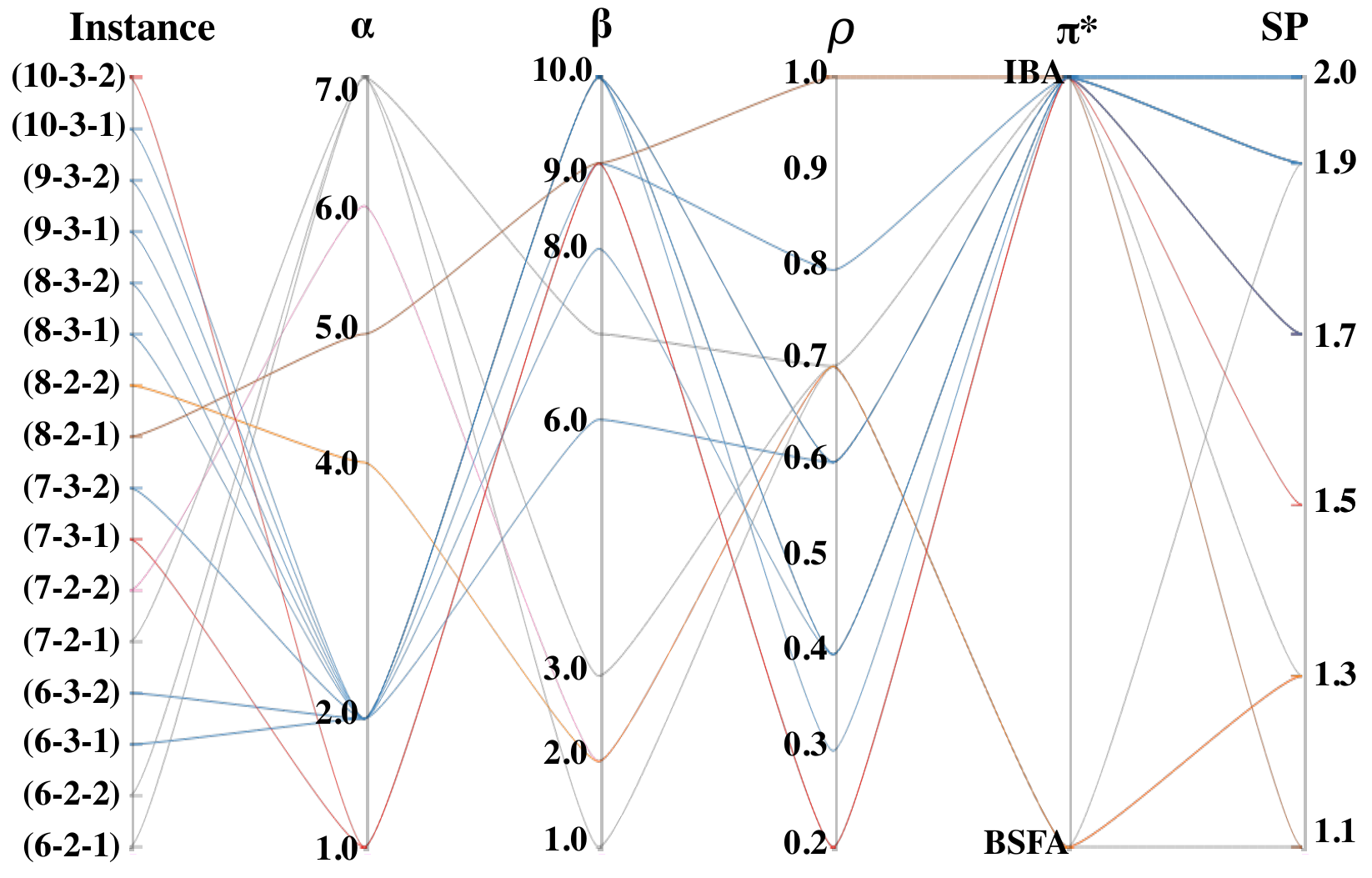}
	\caption{Best parameter configurations for MMAS}
	\label{fig.irace_mmas}
\end{figure}

Figure~\ref{fig.irace_rga} and~\ref{fig.irace_mmas} show the configurations obtained by Irace for each instance for RGA and MMAS, respectively. These parallel coordinate plots illustrate the value of each parameter for an obtained setup which is represented by a line. Note that these plots only show the configuration obtained for instances with more than 6 deliveries.

For RGA, we can see that selection pressures greater than 1.7 are more favourable; the greedy parameter value should be small or large, with no value in the range of 4-6. 10 and 11 have not been identified for tuning.
For MMAS, we can see that selection pressure follows the same pattern in RGA. IBA is the best choice for updating the pheromones for most instances. All values other than 0.1, 0.5, and 0.9 have been utilised for evaporation rate ($\rho$). We can also see that the chosen values for $\alpha$ and $\beta$ are substantially different from ACO's most typical settings of $\alpha=1$ and $\beta=2$.

\section{results}
\setlength{\tabcolsep}{2pt}

\begin{sidewaystable*}[]
\caption{Objective functions obtained for the solutions in for instances with 6-7 deliveries}
\label{table.exp1}
\adjustbox{max width=0.9\textwidth}{}
\begin{tabular}{llllll} 
\toprule 
{} & & DGA & RGA & MMAS & MMAS-local \\ 
Instance & &  & & & \\ 
\midrule 
\multirow{4}{*}{\rotatebox{90}{(6-2-1)}}
& Median & (0.0438,  102.1746, 197.4267) & (0.0046,  104.2092, 246.6884) & (0.0011,  105.0168, 319.2173) & \B (0.001,  99.2849, 313.7267) \\ 
& Best & (0.0438,  102.1746, 197.4267) & (0.0046,  104.2092, 246.6884) & (0.0009,  106.1539, 313.9708) & \B (0.0009,  99.832, 297.235) \\ 
& Worst & (0.0438,  102.1746, 197.4267) & (0.3946,  115.1596, 314.2067) & \B (0.0013,  104.9209, 319.558) & (0.0014,  100.8427, 300.9663) \\ 
& SR (\%) & 0.0 & 0.0 & 0.0 & 0.0 \\ 
\cmidrule{2-6} 
\multirow{4}{*}{\rotatebox{90}{(6-2-2)}}
& Median & (0.0235,  103.1665, 220.0882) & (0.0153,  103.351, 303.5167) & (0.001,  102.6809, 407.1123) & \B(0.0009,  101.173, 366.9152) \\ 
& Best & (0.0235,  103.1665, 220.0882) & (0.0024,  102.9986, 273.2049) & (0.0009,  101.834, 429.2643) & \B(0.0009,  98.2865, 369.7729) \\ 
& Worst & (0.0235,  103.1665, 220.0882) & (0.119,  104.6899, 371.9484) & (0.001,  105.9105, 412.3481) & \B(0.001,  100.9826, 388.3912) \\ 
& SR (\%) & 0.0 & 0.0 & 0.0 & 0.0 \\ 
\cmidrule{2-6} 
\multirow{4}{*}{\rotatebox{90}{(6-3-1)}}
& Median & (0.0,  105.3406, 215.4017) & (0.0,  107.8442, 222.0722) & (0.0,  98.1578, 199.2548) & \B(0.0,  96.0026, 196.5737) \\ 
& Best & (0.0,  105.3406, 215.4017) & (0.0,  101.5879, 216.1834) & (0.0,  96.1019, 205.8208) & \B(0.0,  95.2302, 192.9204) \\ 
& Worst & (0.0,  105.3406, 215.4017) & (0.0,  110.0313, 235.1967) & (0.0,  98.2292, 195.2234) & \B(0.0,  97.9712, 183.2543) \\ 
& SR (\%) & 100.0 & 100.0 & 100.0 & 100.0 \\ 
\cmidrule{2-6} 
\multirow{4}{*}{\rotatebox{90}{(6-3-2)}}
& Median & (0.0,  107.1645, 224.3089) & (0.0,  108.7695, 237.8931) & (0.0,  99.8783, 195.6318) & \B(0.0,  99.6923, 205.196) \\ 
& Best & (0.0,  107.1645, 224.3089) & (0.0,  103.2542, 223.0722) & \B(0.0,  97.6263, 255.3897) & (0.0,  97.7767, 192.1045) \\ 
& Worst & (0.0,  107.1645, 224.3089) & (0.0,  112.1085, 252.1892) & (0.0,  100.622, 219.9266) & \B(0.0,  100.5491, 213.4196) \\ 
& SR (\%) & 100.0 & 100.0 & 100.0 & 100.0 \\ 
\cmidrule{2-6} 
\multirow{4}{*}{\rotatebox{90}{(7-2-1)}}
& Median & (0.0438,  102.1746, 216.317) & (0.0462,  102.5059, 234.9085) & (0.0011,  103.0838, 331.9113) & \B(0.0011,  100.5098, 346.8453) \\ 
& Best & (0.0438,  102.1746, 216.317) & (0.0106,  100.4892, 227.1779) & \B(0.0009,  102.0894, 341.0696) & (0.001,  98.3401, 330.7355) \\ 
& Worst & (0.0438,  102.1746, 216.317) & (0.2418,  113.3867, 321.4938) & \B(0.0012,  103.8423, 350.8519) & (0.0013,  99.0285, 323.713) \\ 
& SR (\%) & 0.0 & 0.0 & 0.0 & 0.0 \\ 
\cmidrule{2-6} 
\multirow{4}{*}{\rotatebox{90}{(7-2-2)}}
& Median & (0.0235,  103.1665, 238.108) & (0.0352,  104.8558, 310.5028) & (0.0009,  107.5879, 437.6702) & \B(0.0009,  100.6736, 394.1645) \\ 
& Best & (0.0235,  103.1665, 238.108) & (0.0024,  103.6803, 323.8809) & (0.0009,  103.1755, 417.8908) & \B(0.0009,  98.3221, 390.2379) \\ 
& Worst & (0.0235,  103.1665, 238.108) & (0.1739,  105.3239, 392.1911) & (0.001,  106.6327, 419.3385) & \B(0.001,  100.6576, 398.1106) \\ 
& SR (\%) & 0.0 & 0.0 & 0.0 & 0.0 \\ 
\cmidrule{2-6} 
\multirow{4}{*}{\rotatebox{90}{(7-3-1)}}
& Median & (0.0,  105.3406, 224.2254) & (0.0026,  105.2915, 246.4175) & (0.0,  98.2164, 222.4849) & \B(0.0,  98.0351, 208.0818) \\ 
& Best & (0.0,  105.3406, 224.2254) & (0.0,  100.399, 282.8226) & (0.0,  95.3277, 210.943) & \B(0.0,  95.2497, 209.0721) \\ 
& Worst & (0.0,  105.3406, 224.2254) & (0.0026,  105.2915, 246.4175) & (0.0,  99.8691, 230.628) & \B(0.0,  99.6428, 214.9185) \\ 
& SR (\%) & 100.0 & 98.0 & 100.0 & 100.0 \\ 
\cmidrule{2-6} 
\multirow{4}{*}{\rotatebox{90}{(7-3-2)}}
& Median & (0.0,  107.1645, 233.1917) & (0.0,  108.2167, 266.023) & (0.0,  99.8828, 234.1908) & \B(0.0,  98.954, 279.1073) \\ 
& Best & (0.0,  107.1645, 233.1917) & (0.0,  103.449, 262.2946) & (0.0,  97.9256, 211.2687) & \B(0.0,  97.1307, 266.3458) \\ 
& Worst & (0.0,  107.1645, 233.1917) & (0.0005,  106.7591, 259.0931) & \B(0.0,  100.7634, 209.6184) & (0.0,  101.4503, 220.1084) \\ 
& SR (\%) & 100.0 & 97.0 & 100.0 & 100.0 \\ 
\bottomrule 
\end{tabular} 

\end{sidewaystable*}

\begin{sidewaystable*}[]
\caption{Objective functions obtained for the solutions in for instances with 8-10 deliveries}
\label{table.exp2}
\adjustbox{max width=0.99\columnwidth}{}
\begin{tabular}{llllll} 
\toprule 
{} & & DGA & RGA & MMAS & MMAS-local \\ 
Instance & &  & & & \\ 
\midrule 
\multirow{4}{*}{\rotatebox{90}{(8-2-1)}}
& Median & (0.0438,  102.1746, 251.1864) & (0.0273,  101.6469, 265.3198) & (0.001,  107.6261, 340.2463) & \B(0.001,  100.4274, 382.1242) \\ 
& Best & (0.0438,  102.1746, 251.1864) & (0.0061,  99.7059, 269.7227) & (0.001,  104.662, 364.4957) & \B(0.001,  99.411, 366.9006) \\ 
& Worst & (0.0438,  102.1746, 251.1864) & (0.3213,  106.2929, 350.9886) & \B(0.0014,  108.9955, 350.0817) & (0.0015,  101.8406, 334.152) \\ 
& SR (\%) & 0.0 & 0.0 & 0.0 & 0.0 \\ 
\cmidrule{2-6} 
\multirow{4}{*}{\rotatebox{90}{(8-2-2)}}
& Median & (0.0235,  103.2428, 263.9633) & (0.051,  102.3833, 279.9597) & (0.001,  103.8465, 438.6782) & \B(0.0009,  98.6669, 419.5654) \\ 
& Best & (0.0235,  103.2428, 263.9633) & (0.0099,  100.5136, 281.1246) & (0.0009,  104.757, 464.8563) & \B(0.0009,  98.4481, 420.6694) \\ 
& Worst & (0.0235,  103.2428, 263.9633) & (0.2667,  104.7104, 399.3313) & (0.0011,  110.8108, 428.5091) & \B(0.001,  101.8508, 410.8488) \\ 
& SR (\%) & 0.0 & 0.0 & 0.0 & 0.0 \\ 
\cmidrule{2-6} 
\multirow{4}{*}{\rotatebox{90}{(8-3-1)}}
& Median & (0.0,  105.3626, 261.8697) & (0.0,  104.4384, 311.0165) & \B(0.0,  97.9676, 285.1839) & (0.0,  97.9992, 248.4489) \\ 
& Best & (0.0,  105.3626, 261.8697) & (0.0,  100.5335, 308.9175) & (0.0,  96.2408, 234.5436) & \B(0.0,  96.2062, 257.1488) \\ 
& Worst & (0.0,  105.3626, 261.8697) & (0.0014,  105.9245, 270.8747) & \B(0.0,  98.1712, 308.6494) & (0.0,  99.0334, 242.9122) \\ 
& SR (\%) & 100.0 & 92.0 & 100.0 & 100.0 \\ 
\cmidrule{2-6} 
\multirow{4}{*}{\rotatebox{90}{(8-3-2)}}
& Median & (0.0,  107.1645, 270.9) & (0.0,  103.0842, 293.0588) & \B(0.0,  98.9443, 259.0835) & (0.0,  99.6151, 237.5469) \\ 
& Best & (0.0,  107.1645, 270.9) & (0.0,  103.0842, 293.0588) & (0.0,  98.1982, 281.7398) & \B(0.0,  97.8378, 263.3097) \\ 
& Worst & (0.0,  107.1645, 270.9) & (0.0026,  108.1778, 299.4416) & \B(0.0,  100.4589, 293.1712) & (0.0,  100.4663, 267.3126) \\ 
& SR (\%) & 100.0 & 91.0 & 100.0 & 100.0 \\ 
\cmidrule{2-6} 
\multirow{4}{*}{\rotatebox{90}{(9-3-1)}}
& Median & (0.0,  105.4089, 302.9124) & (0.0,  106.8992, 287.1712) & (0.0,  97.7961, 281.3341) & \B(0.0,  97.7619, 285.255) \\ 
& Best & (0.0,  105.4089, 302.9124) & (0.0,  100.1781, 331.6884) & (0.0,  96.3332, 292.1141) & \B(0.0,  96.0063, 274.6188) \\ 
& Worst & (0.0,  105.4089, 302.9124) & (0.0131,  103.7835, 356.1299) & (0.0,  98.7486, 270.1143) & \B(0.0,  98.033, 276.9183) \\ 
& SR (\%) & 100.0 & 85.0 & 100.0 & 100.0 \\ 
\cmidrule{2-6} 
\multirow{4}{*}{\rotatebox{90}{(9-3-2)}}
& Median & (0.0,  107.1828, 321.2739) & (0.0,  105.8886, 345.9241) & \B(0.0,  98.8647, 305.6628) & (0.0,  100.4717, 305.733) \\ 
& Best & (0.0,  107.1828, 321.2739) & (0.0,  105.1871, 342.7084) & (0.0,  98.8647, 305.6628) & \B(0.0,  98.6388, 289.2822) \\ 
& Worst & (0.0,  107.1828, 321.2739) & (0.0035,  107.1188, 353.2033) & (0.0,  101.6301, 324.3876) & \B(0.0,  100.6362, 322.0756) \\ 
& SR (\%) & 100.0 & 94.0 & 100.0 & 100.0 \\ 
\cmidrule{2-6} 
\multirow{4}{*}{\rotatebox{90}{(10-3-1)}}
& Median & (0.0,  105.4252, 311.0819) & (0.045,  100.8182, 367.499) & \B(0.0,  97.4539, 303.5063) & (0.0,  98.0039, 291.9887) \\ 
& Best & (0.0,  105.4252, 311.0819) & (0.0,  102.7394, 311.4885) & (0.0,  96.3966, 300.2111) & \B(0.0,  96.1314, 347.2854) \\ 
& Worst & (0.0,  105.4252, 311.0819) & (0.045,  100.8182, 367.499) & \B(0.0,  98.3585, 309.6789) & (0.0,  98.5017, 309.4715) \\ 
& SR (\%) & 100.0 & 72.0 & 100.0 & 100.0 \\ 
\cmidrule{2-6} 
\multirow{4}{*}{\rotatebox{90}{(10-3-2)}}
& Median & (0.0,  107.1865, 339.3049) & (0.0,  109.506, 387.0197) & (0.0,  100.4038, 320.7127) & \B(0.0,  99.2589, 327.916) \\ 
& Best & (0.0,  107.1865, 339.3049) & (0.0,  104.5479, 358.7583) & \B(0.0,  98.3857, 319.6994) & (0.0,  98.8018, 322.0849) \\ 
& Worst & (0.0,  107.1865, 339.3049) & (0.0114,  110.5372, 390.1529) & (0.0,  102.8742, 319.4772) & \B(0.0,  101.8307, 325.0809) \\ 
& SR (\%) & 100.0 & 82.0 & 100.0 & 100.0 \\ 
\bottomrule 
\end{tabular} 

\end{sidewaystable*}
We run RGA and MMAS on each instance 51 times to obtain reliable results and the median could be easily calculated. We use the Kruskal-Wallis test with a 95\% confidence interval to compare randomised algorithms and check if there is a significant difference between them. Next, for pair-wise comparisons, we use the Bonferroni posteriori approach to correct the p-values. We rank the obtained solutions according to our lexicographic objective function, and we use this ranking to perform the statistical comparison.

Tables~\ref{table.exp1} and \ref{table.exp2} list the objective function obtained for optimised solution for planning deliveries with more than 6 deliveries. In reported tables, we list the median, best and worst solutions obtained for our randomised algorithms. We report the same value for DGA because it is deterministic.
 
We use a success rate (SR) indicator to show the percentage of observations for a randomised algorithm where the first component of the objective function (the most important objective) for an obtained solution is zero $v_1(x)=0$. Note that SR for DGA (as a deterministic algorithm) is either 0.0 or 100.0.

\begin{figure*}[t] 
	\centering
	\includegraphics[width=\columnwidth]{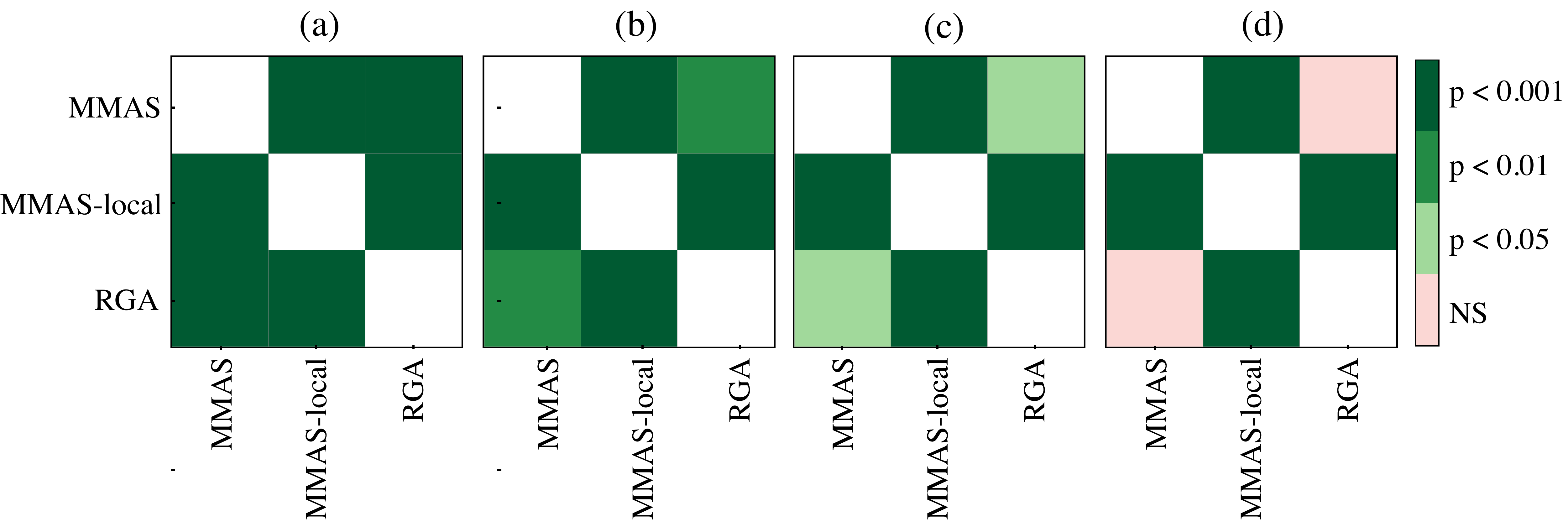}
	\caption{Significance plot of statistical tests for randomised algorithms for different instances. $p$ refers to the p-value and NS shows no significant difference. Each subplot refers to different instances as follows. (a): (3-2-2), (4-2-1), (4-2-2), (4-3-2), (5-3-2), (7-2-2), (9-3-2). (b): (5-2-1), (7-3-2). (c): (6-3-2), (d): other instances.
}
	\label{fig.sig}
\end{figure*}

Figure~\ref{fig.sig} shows the statistical significance for all instances. There are four categories of significant differences observed. In all cases, there is a strongly significant difference between MMAS-local and other algorithms. In some instances, tuned MMAS and RGA are weakly different; however, they are not significantly different in most instances.

DGA is successful in finding solution with 100\% SR in 10 instances out of 16. As observed in~\cite{SAC21}, DGA exhaust good material early in the stages of planning. In all instances, RGA outperforms DGA. We can see the same trend in outperforming RGA by MMAS. 

We can see that planning six deliveries or less using two reclaimers instead of three can lead to a slight violation in $v_1(x)$. However, planning by three reclaimers results in a 100\% success rate for all randomised algorithms.

As the number of delivery surpasses 6, we can see that RGA becomes less capable of finding solutions with 100\% SR. On the other hand, MMAS and MMAS-local can find solutions with 100\% SR always for these instances. MMAS-local can outperform MMAS in most cases, but there exist some exceptions in our observations. We see that for instance (6-3-2), the best solution obtained by MMAS is slightly better than the one obtained by MMAS-local with respect to $v_1(x)$. However, MMAS-local has found a worse solution, but with a better utility.  The median and worst of obtained solutions by MMAS-local is better than corresponding ones obtained by MMAS. We can see the same trend for instances (7-2-1), (10-3-2).

There could be different reasons for this limitation. It could be due to the nature of our local search, in which we just shift the jobs in a solution without changing their reclamation directions. Therefore, there is potential to expand on the local search strategy to consider the reclamation direction in specific where bi-directional reclamation is occurring such as instance (6-3-2).

Another possibility is that there is a trade-off between $v_1(x)$ and $u(x)$. For instance (7-2-1), the difference in $v_1(x)$ between the solutions generated by MMAS-local and MMAS is 0.0001 for the best obtained solution. We can see that MMAS-local has obtained a better solution despite the minor variance in the first component. According to this observation, it is better to define a threshold to have flexibility in dealing with hard constraints on comparing two solutions to identify more practical solutions.

\section{Conclusions}

This work looked at a stockyard recovery problem with multiple reclaimers to schedule short and long-term deliveries while avoiding reclaimers crossing each other. In order to compare the two solutions, we prioritised minimising penalty fees due to contaminants in delivery over operating costs. To simulate the interaction of the reclaimers in constructing a valid solution, we developed a solution construction heuristic. To select reclamation jobs at each step, we investigated deterministic and randomised greedy algorithms, as well as a variation of ant colony optimisation. We also used the automatic parameter tuning method to fine-tune our algorithms and determine the best configuration for each instance to achieve better results. We also designed a local search operator for MMAS to more finely investigate a solution, and it is promising where it can outperform other methods in most instances.

Further research may include (1) Adding a human in the design loop to revise the algorithms' solutions in order to facilitate the practical use of this work, (2) improving the local search to account for changes in the reclamation direction and adding a threshold for trade-off consideration among objectives, and (3) applying the platform to a dynamic problem of stacking and reclaiming a stockpile where in practice stacking occurs in parallel to reclaiming.
\section*{Acknowledgements}
This research has been supported by the SA Government through the PRIF RCP Mining Consortium and by the Australian Research Council discovery project DP200102364 and the training centre IC190100017.


\bibliographystyle{plain}
\bibliography{sample-bibliography}

\end{document}